\documentclass[aps,prb,preprint]{revtex4-1}

\usepackage{graphicx}

\usepackage{amssymb}

\usepackage{amsmath}
\usepackage{verbatim}

\begin{document}

\title{A simulated annealing approach to the student-project allocation problem}



\author{Abigail H. Chown}
\affiliation{Department of Physics, University of Bath, BA2 7AY, United Kingdom}

\author{Christopher J. Cook}
\affiliation{Department of Physics, University of Bath, BA2 7AY, United Kingdom}

\author{Nigel B. Wilding}
\email{n.b.wilding@bath.ac.uk}
\affiliation{Department of Physics, University of Bath, BA2 7AY, United Kingdom}


\begin{abstract}
We describe a solution to the student-project allocation problem using simulated annealing. The problem involves assigning students to projects, where each student has ranked a fixed number of projects in order of preference. Each project is offered by a specific supervisor (or supervisors), and the goal is to find an optimal matching of students to projects taking into account the students' preferences, the constraint that only one student can be assigned to a given project, and the constraint that supervisors have a maximum workload. We show that when applied to a real dataset from a university physics department, simulated annealing allows the rapid determination of high quality solutions to this allocation problem. The quality of the solution is quantified by a satisfaction metric derived from empirical student survey data. Our approach provides high quality allocations in a matter of minutes that are as good as those found previously by the course organizer using a laborious trial-and-error approach. We investigate how the quality of the allocation is affected by the ratio of the number of projects offered to the number of students and the number of projects ranked by each student. We briefly discuss how our approach can be generalized to include other types of constraints and discuss its potential applicability to wider allocation problems.

\end{abstract}



\maketitle

\section{Introduction and background}
\label{S:1}
The student-project allocation problem is encountered at the majority of British universities. Students are typically required to undergo a supervised project, often during their final year of study, in a field of a subject that interests them. There is a list of projects, each submitted by a project supervisor, and students are required to rank a certain number of projects in order of preference. This process is the one-sided variant of the problem.

Usually only one student works on a given project. Thus, issues arise when several students choose a particular project as their first choice preference. Inevitably, all but one of these students will be assigned one of their less preferable projects, which in turn causes secondary effects because the less preferred projects could be the first choice of another student. An additional issue faced is the supervisor's workload. Usually a supervisor proposes more projects than they can feasibly oversee in order to provide a greater range of student choices. Hence, further complications arise when all of a particular supervisor's projects are popular because it guarantees some students will not be assigned their first choice project.

The student-project allocation problem is a specific case of the generalized assignment problem, a well-known optimization problem that consists of assigning sets of jobs to sets of agents while minimizing the cost associated with the assignment; this problem can be formulated as an integer linear problem.\cite{Osman:1995} A wide range of methods have been devised for solving such problems including, but not limited to, a genetic algorithm\cite{CHU199717} and a local-ratio technique for the knapsack problem.\cite{COHEN2006162}

There have been different approaches to solving the one-sided student-project allocation problem. (The problem is one-sided because supervisors do not rank the students.) Proll used a bottleneck-assignment method and showed that this method is superior to a standard first-in first-out method.\cite{Proll1972} Anwar and Bahaj used a two-stage integer program, first attempting to reduce supervisor workload, before then looking at student satisfaction.\cite{Anwar2003} Another approach\cite{Harper2005} used a genetic algorithm. More recently, Kwanashie {\em et al.}\cite{Kwanashie2015} modeled the student-project allocation as a network flow problem, and also investigated different matching criteria.

There has also been significant research into solving the two-sided student-project allocation problem, which is a generalization of the hospital-residents problem.\cite{ABRAHAM200773,MANLOVE2008553} Abraham {\em et al.}\ developed two algorithms for the problem, one student-oriented and one supervisor-oriented.\cite{ABRAHAM200773} However, these algorithms appear not to have been tested in practice. Manlove and O'Malley\cite{MANLOVE2008553} investigated a variant, where supervisors had preference over projects and not students. They showed it was NP-hard, and presented an approximation algorithm.

In this paper we investigate how simulated annealing, a method from statistical and computational physics, can be used to obtain a good solution to the student-project allocation problem. We demonstrate the utility of our approach on real student data and provide access to our code and datasets so that others can experiment with the methodology.

\section{Problem specification}
\label{sec:probspec}
We focus on the assignment of final year projects at the Department of Physics of the University of Bath. With minor modifications, our method should be applicable to a wide range of related assignment problems.

Each student at Bath is invited to submit a list of four project preferences, ranking them from 1, their most preferred project, to 4, their least preferred. The course organizer then assigns students to projects with the goal of maximizing student satisfaction, while satisfying the following constraints:
\begin{enumerate}
\item Each student is assigned to one of the projects on their preference list.
\item No project can be assigned to more than one student.
\item Supervisors cannot be assigned more projects than they can feasibly oversee.
\end{enumerate}

The problem can be framed mathematically. Let $N$ be the number of students, $M$ the number of projects (with $M>N$), and $S$ the number of supervisors. Define a $N\times M$ matrix $C$ with elements
\begin{equation}
C_{ij} = \begin{cases}
1 & \text{if student $i$ chose project $j$} \\
0 & \text{otherwise,}
\end{cases}
\end{equation}
and define an $N\times M$ allocation matrix $X$ by
\begin{equation}
X_{ij}= \begin{cases}
1 & \text{if student $i$ is assigned to project $j$} \\
0 & \text{otherwise.}
\end{cases}
\end{equation}

Our goal is to maximize student satisfaction. To do so, we need a definition of satisfaction. Clearly students will be happier if they receive their first choice and less happy if they receive their fourth choice. We define
\begin{equation}
\label{eq:objfun}
{\mathcal O}=\sum_{k=1}^{4} w_{k}n_{k}\:,
\end{equation}
where $n_k$ is the number of students assigned their $k$th choice in the allocation and $w_{k}$ is the weighting assigned to the $k$th choice, with $w_{k}>w_{k+1}$. The sum in Eq.~(\ref{eq:objfun}) is our measure of overall class satisfaction. In the optimization literature such a quantity is referred to as the {\em objective function}.

The first two constraints can be written simply as
\begin{align}
\sum_{j=1}^{M} C_{ij}X_{ij} & = 1 \hspace{0.5cm} \forall\, i = 1,\ldots,N
\label{eq:con1} \\
\sum_{i=1}^{N} X_{ij} & \leqslant 1 \hspace{0.5cm} \forall\, j = 1,\ldots,M.
\label{eq:con2}
\end{align}

To incorporate the third constraint, we let $F_{s}$ denote the number of projects that supervisor $s$ has in the assignment, and $L_{s}$ denote the largest number of projects that they can oversee. Then
\begin{equation}
F_{s} \leqslant L_{s} \hspace{0.5cm} \forall\, s = 1,\ldots,S.
\label{eq:con3}
\end{equation}
Given a matrix of choices $C$, our goal is to find an allocation $X_{ij}$ that maximizes the value of ${\mathcal O}$ given the weights $w_{k}$ and the constraints in Eqs.~(\ref{eq:con1})--(\ref{eq:con3}). In the following we describe a simulated annealing algorithm that achieves this goal.

\section{Method}
\label{sec:method}
\subsection{Monte Carlo simulation}
Simulated annealing is a stochastic method for finding the global minimum of a function. It is particularly useful if the function is badly behaved, for example, if it has many local minima. Simulated annealing is an adaptation of the Monte Carlo simulation technique which is used to generate arrangements of the constituents of a physical system with the correct probability. For a system of particles in thermal equilibrium with its surroundings, this probability corresponds to the Boltzmann distribution:
\begin{equation}
\label{eq:boltz}
P({\bf a}) = \frac{1}{Z} \sum_{{\bf a}} e^{-E({\bf a})/(k_BT)}.
\end{equation}
The label ${\bf a}$ represents all possible arrangements of the particles, and the energy of a particular arrangement is $E({\bf a})$. The constant $k_B$ sets the energy scale, and $T$ is the temperature. For a particular temperature, $Z$ is a constant that serves to normalize the probability distribution.

Suppose we are interested in an observable $Q$ whose value depends on the particular arrangement ${\bf a}$. The average value of $Q$ is given by
\begin{equation}
\label{average}
\langle Q \rangle = \sum_{{\bf a}}P({\bf a}) Q({\bf a}).
\end{equation}
Monte Carlo methods allow us to estimate $\langle Q \rangle$ by approximating the sum in Eq.~(\ref{average}), which is over all arrangements, by a sum over a subset of arrangements. This estimate is achieved by sampling; that is, many different particle arrangements are generated randomly, and the value of $Q$ is averaged over the subset of sampled arrangements. The larger this subset, the closer the estimate for $\langle Q \rangle$ is to its true value. Usually, such simple sampling is inefficient and imprecise, and smarter sampling methods are needed. One such method is the Metropolis algorithm.

\subsection{Metropolis Algorithm}
The Metropolis algorithm implements a filter that ensures that only those arrangements that contribute most to the sum in Eq.~(\ref{average}) are sampled. The filter is known as importance sampling and operates via a Markov process as follows.\cite{Binder:2010} Starting from some arrangement, we use a pseudorandom number generator to propose a trial arrangement that differs from the current one in some way. The trial arrangement is accepted as the new arrangement with some probability. Suppose that the current arrangement has energy $E({\bf a})$, and the trial arrangement has energy $E({\bf a}^\prime)$. Then the acceptance probability is
\begin{equation}
\label{probabilitycondition}
P({\bf a} \rightarrow {\bf a}^\prime) = \frac{P({\bf a}^\prime)}{P({\bf a})} = e^{-\Delta E/(k_bT)},
\end{equation}
where
\begin{equation}
\label{deltaE}
\Delta E = E({\bf a}^\prime) - E({\bf a}).
\end{equation}
As a result, only the energies of the arrangements, which are known values, are needed to calculate the transition probability and the constant $Z$ cancels and can be ignored.

In practice, the Metropolis algorithm operates as follows:
\begin{enumerate}
\item Choose an initial arrangement ${\bf a}$.

\item Choose a trial arrangement ${\bf a}^\prime$.

\item Calculate the energy difference in Eq.~(\ref{deltaE}).

\item Generate a random number $r \in (0,1)$.

\item If $r < P( ${\bf a}$\rightarrow ${\bf a}$^\prime)$, move to state ${\bf a}^\prime$.

\item Generate a new trial arrangement and return to step~3.
\end{enumerate}

\begin{figure}[t!]
\centering
\includegraphics[height=0.25\textheight]{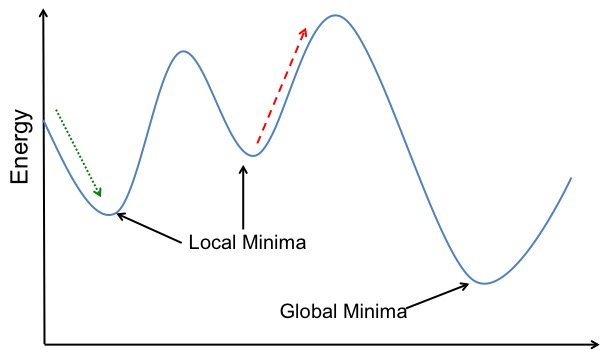}
\caption{Illustration of how the Metropolis algorithm is used to explore the space of arrangements. The dotted arrow to the left of the graph depicts downhill searching and the dashed arrow towards the right of the graph represents uphill searching. (Color online)}
\label{fig:Metropolis}
\end{figure}

The Metropolis algorithm allows the system to explore its arrangements with a probability proportional to their Boltzmann weight. The process can be visualized by constructing a one-dimensional graphical representation of the space of arrangements, parameterized by the system energy $E$, as shown schematically in Fig.~\ref{fig:Metropolis}. In the Metropolis algorithm, downhill moves (ones that reduce the energy) are always accepted, as shown by the green, dotted arrow in Fig.~\ref{fig:Metropolis}. In contrast, uphill moves (ones that increase the system energy as shown by the red, dashed arrow in Fig.~ \ref{fig:Metropolis}) are exponentially suppressed and accepted with a probability $e^{-\Delta E/k_bT}$. The possibility of uphill movement allows the sample to escape local minima in the space of arrangements, provided the depth of the minimum is not much greater than $k_bT$. It is this feature of the Metropolis algorithm which is exploited by simulated annealing to increase the probability of finding the global minimum of a function.

\subsection{Simulated Annealing}
Simulated annealing is a stochastic method for finding the global minimum of a function. It is inspired by techniques in materials science. If a liquid is cooled slowly, it will form a pure crystal, corresponding to the lowest energy state. However, if it is cooled quickly, it will end up in an amorphous state with higher energy. Simulated annealing works by marrying this concept to Monte Carlo methods through the Metropolis algorithm.

The general idea is to initiate the simulation at a high temperature $T$, and slowly decrease $T$ as the simulation progresses. The initial high temperature allows the system to explore energetically unfavorable arrangements, but the frequency with which these occur decrease as the system cools. In principle, this procedure should allow the system to escape local energy minima and increase the chance of finding the global minimum.

Although there is debate \cite{Clemons2004} as to how appropriate it is to compare the physical annealing process to that of combinatorial optimization, this approach naturally lends itself to optimization. If we consider the space of possible allocations to be the phase space and the negative of the objective function in Eq.~(\ref{eq:objfun}) to play the role of energy (noting that maximization problems and minimization problems are interchangeable), then we can use simulated annealing to find an optimal solution to the student-project allocation problem.

\section{Implementation for the Student-project Allocation problem}
\label{S:4}
We can find a good solution of the student-project allocation problem using the Metropolis algorithm and simulated annealing techniques. To this end we have created a computer program written in the C programming language that is available for download from GitHub and AJP's supplemental material server, see appendix).  We will not discuss all the details of how the program operates, but will cover some key points.

\subsection{Objective Function}
We are aiming to maximize student happiness and hence define our energy in terms of the objective function (\ref{eq:objfun}) as
\begin{equation}
E = -\mathcal {O}
= -w_{1}n_{1} -w_{2}n_{2} -w_{3}n_{3} -w_{4}n_{4}.
\label{eq:SAobjfun}
\end{equation}
Recall that $w_{k}$ is the weight of preference $k$ and $n_{k}$ is the number of pairs allocated their $k$th preference. Naturally, the choice of weights should decrease with $k$; that is, a first choice should receive a greater weight than the second choice, etc. We initially choose to work with a set of linearly decreasing weights, leading to the objective function
\begin{equation}
\label{eq:linwhts}
E = -4n_{1} -3n_{2} -2n_{3} -n_{4}\:.
\end{equation}

This objective function suffices when tackling a particular student-project allocation dataset. However, it does not allow us to compare the quality of solutions among different datasets, which will usually consist of different numbers of student cohorts $N$. Thus, it is useful to define a normalized objective function $E^\star$ where

\begin{equation}
\label{normObj}
E^\star = -\frac{100}{N} \big[ n_{1} + \frac{3}{4}n_{2} + \frac{1}{2}n_{3} + \frac{1}{4}n_{4} \big].
\end{equation}

Using this objective function, the energy corresponding to the best possible allocation has a value of $E^\star=-100$. The best possible allocation is where every student gets their first choice project, 

\subsection{Sampling strategy}

The starting point for a simulated annealing optimization must be an initial allocation that satisfies the constraints in Eqs.~(\ref{eq:con1})--(\ref{eq:con3}). To obtain a suitable initialization we randomly assign each student one of their four choices. In general, this assignment will lead to multiple constraint violations with different students being assigned the same project and supervisors being overworked. To correct these violations, we perform a preliminary Monte Carlo procedure in which we choose a student at random and randomly re-allocate them to one of their four project choices. If this proposed change results in a reduction in the number of violated constraints, we accept the move; otherwise, we reject it. This process is iterated until an allocation is obtained with no violations, which we take as our initial configuration for the application of our simulated annealing algorithm.

An important aspect of the simulated annealing procedure is the manner in which trial allocations are proposed within the Markov process. The method adopted should be reversible, that is, satisfy detailed balance,\cite{Binder:2010} and ergodic -- all allocations should be reachable within the sampling procedure. These conditions can be achieved by iterating the following procedure:
\begin{enumerate}
\item Choose a student at random.

\item Choose a random integer from 1 to 4.

\item If the student is currently assigned the project corresponding to that choice, repeat step~2. Otherwise, assign the student the trial project corresponding to the random choice. This assignment is the trial allocation.

\item Check that for the trial allocation no project is assigned to multiple pairs. If a conflict has arisen, retain the current allocation and go to step~1.

\item Check that the supervisor workload limit is not violated. If it is, then retain the current allocation and go to step~1.

\item Calculate the Metropolis acceptance probability $P({\bf a} \rightarrow {\bf a}^\prime)$ using Eq.~(\ref{probabilitycondition}). Generate a random number $r\in (0,1)$. If $P < r$, retain the current allocation and go to step~1. If $P \ge r$. accept the trial allocation as the current allocation.

\end{enumerate}
Note that because we chose both the student and project randomly for the trial allocation, this procedure ensures ergodicity and is trivially reversible. Also note that because step~4 is more likely to reveal a conflict than step~5, it is computationally expedient to order them as listed.

\subsection{Annealing schedule}
\begin{figure}[t!]
\centering
\includegraphics[height=0.25\textheight]{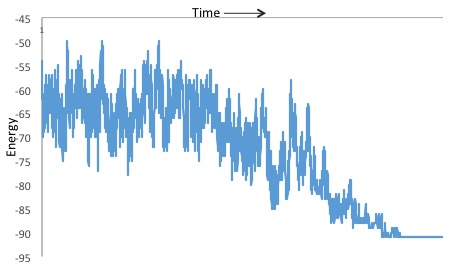}
\caption{Time series for a typical simulated annealing run using dataset $D_{4}$ as input. Note the steady decrease of the value of the energy under the cooling schedule described in the text.}
\label{fig:TimeSeries}
\end{figure}

The annealing schedule is crucial to implementing simulated annealing and specifies how many Monte Carlo moves are made for a given temperature, the start temperature, the end temperature, and the temperature step size. One Monte Carlo move means to carry out steps 1 through 6 from the previous section. Both successful moves and unsuccessful moves contribute to the total number of Monte Carlo moves. These choices are important because if we misjudge the annealing schedule, the system might not find the global minimum or take too long to do so. By trial and error, we found that an initial temperature $T/k_B=5$, final temperature $T/k_B=0$, and temperature step size of $\Delta(T/k_B)=0.001$ worked well for all datasets. In addition, we found that the number of Monte Carlo moves performed at a fixed temperature should be the larger of $1000N$ attempted moves and $100N$ successful moves ($N$ is the number of students). These choices strike a balance between the quality of optimization and the required computational time. An example time series showing how the energy of the allocation decreases under the annealing schedule is shown in Fig.~\ref{fig:TimeSeries}.

\section{Results}
\label{sec:results}

In the following we describe the application of our simulated annealing program to real datasets for student project preferences. In each case we compare the optimization by simulated annealing to that found previously by the course organizer via a laborious trial and error approach.

\begin{table}[th!]
\centering
\begin{tabular}{| c | c | c | c |}
\hline
Dataset & Students, N & Projects, M & Supervisors, S \\ \hline \hline
D1      & 19          & 58          & 27             \\ \hline
D2      & 28          & 58          & 25             \\ \hline
D3      & 24          & 67          & 30             \\ \hline
D4      & 26          & 75          & 30             \\ \hline
\end{tabular}
\caption{Class sizes, number of projects offered and number of supervisors for each dataset}\label{tab:a}
\end{table}

\begin{table*}[bh!]
\label{bigTable}
\centering
\begin{tabular}{ | p{5.6cm} | c | c | c | c |}
\hline
Dataset & $D_1$ & $D_2$ & $D_3$ & $D_4$ \\
\hline\hline
organizer's allocation $E^\star$ & $-86.84$ & $-81.25$ & $-86.46$ & $-86.54$ \\ \hline
simulated annealing minimum $E^\star$ & $-86.84$ & $-82.14$ & $-86.46$ & $-87.50$ \\ \hline
simulated annealing average $E^\star$ & $-86.78 \pm 0.29$ & $-81.52 \pm 1.79$ & $-85.83 \pm 0.62$ & $-87.45 \pm 0.22$\\ \hline
\end{tabular}
\caption{The minimum, average values, and standard deviation of the objective function $E^\star$ resulting from $20$ independent simulated annealing runs on four distinct historical datasets. For comparison, the energy of the solution previously obtained by the course organizer is given.}\label{tab:b}
\end{table*}

\subsection{Comparison via the objective function}

Four datasets were examined and are labelled $D_1$ to $D_4$, corresponding to four years over which the Final Year Project course has offered at the University of Bath. Each dataset varies in class size, total number of projects offered and number of potential supervisors, with values given in Table \ref{tab:a}. To determine the performance of our simulated annealing approach, we ran the algorithm twenty times on each dataset. Each run employed a different sequence of random numbers. A key observable was the final value of the objective function, that is, the energy $E^\star$, Eq.~(\ref{normObj}). Due to the stochastic nature of simulated annealing and the presence of many solutions whose energy is close to the global minimum, the simulation does not yield the same solution each time it is run, nor does it necessarily find solutions with the same energy. Table~\ref{tab:b} gives the minimum value of $E^\star$ found in the twenty runs and the average value and the standard deviation. Also shown is the value of $E^\star$ for the allocation found previously by the course organizer.

We found that for $D_1$ and $D_3$ the allocation made by the course organizer has the same energy as found by the simulated annealing program. For $D_2$ and $D_{4}$ the program finds a slightly better solution than was determined by hand. Interestingly, the variance of the solution energy is much larger for $D_2$ and $D_3$. The failure of simulated annealing to always find a minimal solution is not ideal. It suggests that the energy landscape (see Fig.~\ref{fig:Metropolis}) is rough and that we may have to further investigate the settings of our annealing schedule. However, it is encouraging that even when the program returns a solution that is less energetically favorable than the allocation found by hand, it still is a good solution to the problem.

\subsection{The allocation histogram and degeneracy}
\begin{figure}[t!]
\centering
\includegraphics[height=0.3\textheight]{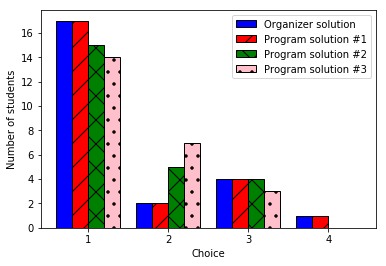}
\caption{Four allocation profiles for dataset $D_3$. The solid blue solution was found by the organizer and red, green, and pink solutions with various markings were found by the simulated annealing algorithm. Each allocation has energy $E^\star= -86.46$. }
\label{fig:D3degeneracy}
\end{figure}

The energy objective function provides a useful, albeit crude measure for comparing allocations. Greater insight into the nature of allocations follows from the allocation histogram, that is, the histogram of numbers of students who received their first, second, third, and fourth choice. For example, consider the case of dataset $D_3$. Figure~\ref{fig:D3degeneracy} shows the allocation histogram for four solutions (three simulated annealing solutions and the course organizer's allocation) which all share the same (minimum) energy $E^\star=-86.46$; that is, they are ``degenerate'' in the energy. We see that the nature of the allocation solutions varies greatly in each case, and two are very different from the course organizer's allocation and one is identical.

This issue of degeneracy of solutions muddies the waters with regard to what constitutes an optimal allocation. Specifically, in the set of allocations shown in Fig.~\ref{fig:D3degeneracy} there are allocations that vary from ``generous'' to ``greedy'' in the terminology of Ref.~\onlinecite{Kwanashie2015}. A greedy allocation is one that aims to maximize the number of first choices allocated, with a disregard for students who receive their fourth choice if it means another student can obtain their first choice. In contrast, a generous allocation makes the overall student population as satisfied as possible and does not allocate a student their first choice if it causes another student to obtain their fourth. To generate allocations of the highest quality, we therefore require additional information regarding how the satisfaction of real students is impacted by whether they received their first, second, third, or fourth choice project. In view of this requirement, our focus in the following is on the choice of weights $w_k$ appearing in Eq.~(\ref{eq:SAobjfun}).

\subsection{Opinion based weightings}

The initial tests of our simulated annealing approach assumed a simple linear decrease of the weights in Eq.~(\ref{eq:linwhts}). However it is unclear whether these weights accurately reflect students' relative satisfaction with the allocations they may be assigned. Furthermore, the use of integer weights in Eq.~(\ref{eq:linwhts}) compounds the incidence of degenerate solutions, and it would be useful to have a means of discriminating between them.

To try to resolve these issues we have created a model based on student opinion. The current cohort of project students was polled to determine how their satisfaction would have varied had they been assigned their first choice, second choice and so on. Students rated each outcome on a scale from $1$ (``very unhappy'') to $5$ (``very happy''). The response rate was 43\% from a class size of 46. Using this data we were able to determine new opinion-based weights: $w_1=4.7, w_2=4.15, w_3=3.0$, and $w_4=2.35$, thus obtaining the new normalized objective function $E^{\prime}$
\begin{equation}
\label{studObj}
E^{\prime} = -\frac{100}{N} \left[ n_{1} + \frac{4.15}{4.7}n_{2} + \frac{3.0}{4.7}n_{3} + \frac{2.35}{4.7}n_{4} \right].
\end{equation}
These weightings are no longer linear in the rank $k$; the difference
between the top two choices and between the bottom two choices is
similar, and there is a larger difference between the weights for the
second and third choice projects. This finding suggests that most
students are still very happy if they receive their second choice
because this choice is now worth 88\% of their first choice, compared
to the value of $75\%$ assumed previously. The third choice is now
worth $63\%$ of a first choice compared to $50\%$ before and a fourth
choice is worth $50\%$ of a first choice compared to $25\%$
previously. All of these ratios have increased, meaning that students
are happier with their lower choices than was initially assumed. In
hindsight, this is not unreasonable given that students choose only
four projects from a large list of approximately $70$ possibilities.

\begin{figure}[t!]
\centering
\includegraphics[height=0.1\textheight]{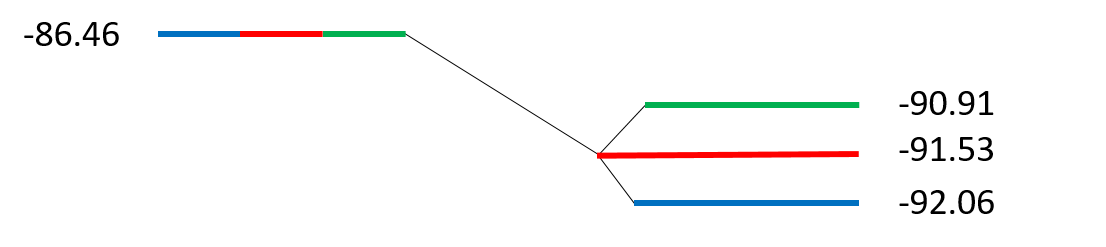}
\caption{Schematic of how the use of opinion based weights applied to dataset $D_3$ lifts the degeneracy of three equal-energy solutions previously found with the linearly decreasing weights.}
\label{fig:degeneracy}
\end{figure}

The new weights engender a shift in the energy scale as might be
expected. For example, when applied to dataset $D_1$ the minimum
energy changed from $E^\star= -86.84$ to $E^\prime= -92.50$ although
in this case (as well as for $D_2$), the actual optimal allocation
remained the same. In contrast, differences in the allocation solution
were found for $D_3$ and $D_4$. With the original weights, simulated
annealing runs on $D_3$ yielded three separate minimum energy
allocations, one of which the course organizer found by hand, all
having the same energy as in Fig.~\ref{fig:D3degeneracy}. Using
non-integer opinion based weights lifts this degeneracy as shown in
Fig.~\ref{fig:degeneracy}, so that now one allocation is clearly
preferable; that is, $E^\star = -92.06$.

Due to the nonlinear dependence of the new weights on the rank choice,
it appears that the new weightings lead to allocations that reduce the
number of third and fourth choices. This presumably occurs because of
the larger relative difference in weights between second and third
choices compared to the original weights.  We also note differences in
the distribution of the energies of the solutions found by the twenty
independent runs. Although the fraction of the twenty runs on datasets
$D_1$, $D_2$, and $D_3$ that yielded the minimum energy solution
increased with the new weights, it decreased for $D_4$: only fourteen
runs yielded the minimum energy compared to nineteen with the old
weights. Nevertheless, the range of solution energies narrowed, with
only two distinct energy solutions found with energies
$E^\prime=-93.21$ and $E^\prime =-93.17$. Thus, although the change in
the objective function initially looks to have adversely affected the
solution of dataset $D_4$, it actually provided more consistent
results for all twenty runs. We also note that it removes all
occurrences of fourth choices in the final allocation of $D_4$, which
were present with the original linearly decreasing weights.

Overall, we find that use of opinion-based weights leads to an improvement in the allocations obtained by simulated annealing. Not only are they based on empirical evidence rather than assumption, but they differentiate previously degenerate solutions and minimize the number of fourth choices. As a result, these weightings will be used by the course organizer for performing the allocation in future years. (However, unless explicitly stated, we will use the original weightings for the rest of the paper.)

\subsection{Effects of tightening the constraints}
\label{sec:tighten}
We now investigate how other model parameters affect the solutions that we find -- specifically the role of the fourth choice and the impact of altering the project-student ratio $M/N$. The only notable previous investigations of the importance of these parameters has been by Kwanashie et al.,\cite{Kwanashie2015} who generated many random datasets and looked at how varying the number of students and the number of choices affected the results for their group project model, which allows multiple students to be assigned to the same project.

There has been limited discussions about the number of choices needed to ensure a successful allocation. Proll\cite{Proll1972} found that for his method a successful allocation required the number of choices $R$ to obey $R \geq \max \{5, 0.2M \}$. Despite having datasets of a similar size to Proll, we do not follow this requirement because our method gives successful solutions with just four choices.

Having established that four choices will yield successful solutions, it is reasonable to ask whether three choices also does so. We first consider $D_2$, whose final allocation always contains a fourth choice. Can we find a solution to $D_2$ without the fourth choices being available? We removed the fourth choices from $D_{2}$ and ran our simulated annealing program twenty times and were not surprised to discover that we could not find any feasible solutions. As expected, $D_{2}$ depends on four choices to find a solution.

\begin{figure}[t]
\centering
\includegraphics[height=0.28\textheight]{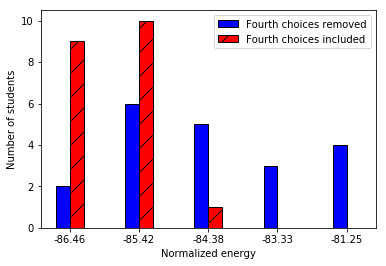}
\caption{Histogram of the energy of simulated annealing solutions obtained from twenty runs on dataset $D_{3}$ with fourth choices included (red diagonal markings) and fourth choices removed (blue solid). Note the wider spread of energies in the case without a fourth choice.}
\label{fig:fourthchoices}
\end{figure}

We also investigated the role the fourth choice plays in the optimization processes for those datasets that usually have final allocations that do not contain any fourth choices. Do they play a role in helping the simulated annealing algorithm find minimum energy configurations or are they disposable? Again we removed the fourth choice and ran the simulated annealing program twenty times for the other three datasets. Datasets $D_{1}$ and $D_{4}$ showed no real statistical differences (as seen before, these datasets are generally easiest to solve). However, we found that the average normalized energy for $D_3$ increases from $-86.46 \pm 0.62$ to $-84.11 \pm 1.72$. The larger standard deviation is a result of the fragmentation of our solutions, as seen in Fig.~\ref{fig:fourthchoices}. There is a wider spread of energies when the fourth choice is removed, which suggests that the fourth choice is used to help move solutions toward the minimal end of the energy scale.

We conclude that four choices are valuable for the simulated annealing optimization procedure. Even if the final allocation does not contain them, they serve as a useful stepping stone. The reason for this dependence on fourth choices can be understood from the point of view of optimization theory, where it is common to introduce more variables than needed to assure that the problem becomes better behaved. One parameter to consider in the future is the introduction of a fifth choice. Although it does not seem necessary currently --- and students would not be happy if they received a fifth choice --- it might make the system easier to solve in challenging situations such as $D_{2}$.

One of the main factors that determine how ``hard'' it is to solve the student-project allocation problem is the ratio of projects to students. It might be expected that the greater the number of projects for students to choose from, the more first choices will be assigned because there is less overlap of student choices. Because the number of available supervisors is roughly constant each year, adding more projects means these supervisors would need to propose more. However, the same restrictions will apply with regards to the supervisor workload and so if all the projects offered by some particular supervisor are popular, then a greater number of students will be disappointed if this supervisor put forward more projects. We look to find an optimal balance that delivers sufficient student choices without too many projects being offered per supervisor.

We take our existing datasets and alter them to change the project-to-student ratio. We could either change the number of projects, keeping student numbers fixed, or alter the number of students, keeping the number of projects fixed. We choose the latter because it is more realistic in practice to change the number of students because the course organizer has no control over the number of projects. To increase the ratio we removed students at random. To decrease the ratio we added students and assign them four project choices chosen randomly. We then considered project-to-student ratios from 1.5 to 1 to 4 to 1, in intervals of $0.5$, and ran each ratio ten times for each dataset.

\begin{figure}[t!]
\centering
\includegraphics[height=0.3\textheight]{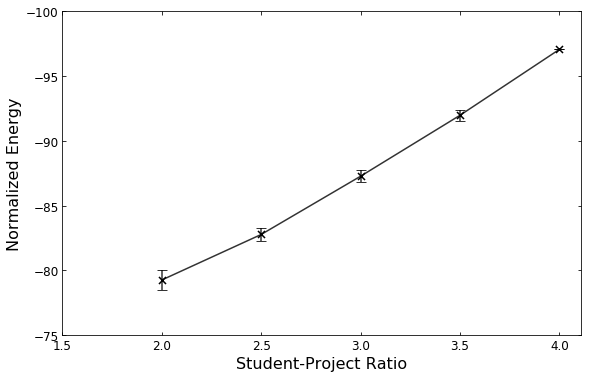}
\caption{Average energy (from ten runs of the simulated annealing program) as a function of the project-student ratio for dataset $D_{3}$. Error bars are the standard deviation at each ratio.}
\label{fig:ratio}
\end{figure}

We noticed immediately that increasing the project-student ratio results in solutions with lower energies, which means that the solutions are converging to the ideal state in which all students are allocated their first choice. One particularly desirable result comes from $D_{3}$. Not only does its energy show an almost linear dependence on the ratio as seen in Fig.~\ref{fig:ratio}, but it also yields the best solution that we have observed in any of our investigations -- a ratio of 4 to 1, with energy $E^\star=-97.06$ for all ten runs of the simulated annealing program. As expected, higher ratios also correspond to more first choices in a configuration. We find that for ratios of 3.5 to 1 and 4 to 1, no solutions contain fourth choices -- not even those for the troublesome $D_{2}$ dataset. In contrast, for a ratio of 1.5 to 1 we did not find any feasible solutions for $D_{1}$ and $D_{3}$, but were able to do so for $D_{2}$ and $D_{4}$. This result is not surprising. Our model cannot assign students to projects outside of their preference list. When the ratio of projects to students is low, there will be significant overlap in the choices made by each student. In general, when we decrease the ratios and consider the 2.5 to 1 and 2 to 1 cases, the quality of our solutions deteriorates, and we have higher energies, fewer first choices and more fourth choices.

These results are useful, but should be treated with some caution. To increase the project-student ratio beyond its true value, students were removed from the dataset at random. To confirm our results we should generate multiple datasets for the same ratio --- which would all be different due to the randomness --- and see if the results vary depending on which students are removed. We also need to be cautious about the results for decreasing the project-student ratio, where we added fictitious students. Their project ``choices'' were chosen randomly, which fails to take into account project popularity. An alternative approach would be to ask different sized groups of students to choose preferences from the same list of projects.

\section{Discussion}
\label{sec:concs}
From a comparison of our results to those obtained by the course organizer using trial-and-error, we see that simulated annealing replicates or improves the solutions obtained in the vast majority of cases, and requires much less of the course organizer's time. The few cases where we find poorer solutions still represent high quality allocations which may be improved further by refining the annealing schedule used.

We found that at least four project choices per student are necessary to obtain good solutions of the standard cases of student-project allocation and that having a higher ratio of projects to students improves the quality of the solutions found. There is room for further investigations using more datasets than we had at our disposal. We suggest that future investigations into these properties follow the lead of Ref.~\onlinecite{Kwanashie2015} and generate many random datasets to enable more statistically reliable conclusions.

Although direct comparisons are difficult, our approach performs favorably with regard to the proportion of first choices in an allocation. Unlike previous efforts, such as Teo and Ho\cite{Teo1998} whose allocations are likely to leave a subset of students unassigned to a project, we always find solutions where every student is assigned to one of their choices. However, our model has a longer runtime, and it is not clear how it will scale to the larger problems that other universities, such as the ones used by Teo and Ho, may face. It would be interesting to investigate this aspect in further work.

Although simulated annealing can be proved to be convergent for
logarithmic cooling schedules \cite{Geman1984}, convergence to the
optimal allocation in polynomial time cannot be guaranteed. In
contrast, methods such as minimum-cost network flow \cite{Kwanashie2015}
can be shown to deliver the optimal allocation in polynomial time. To
make the simulated annealing approach competitive in terms of wall
clock time, one speeds up convergence by utilizing heuristic fast
cooling schedules as we have done, but this speed up is achieved at
the cost of possibly missing the optimal solution. Nevertheless, we
argue that this tradeoff is reasonable for the student-project
allocation problem, because one is usually looking for a ``good
enough'' allocation, not necessarily the best one. We note that there
are reports\cite{Kwanashie2015} that the minimum-cost network flow
approach suffers numerical issues if the weights are set to favor
greedy allocations.

Our simulated annealing program could have many other applications. It could be adapted to other one-sided assignment problems such as allocating primary school students to one of their chosen secondary schools. The objective function could be adapted to include factors affecting this allocation such as the distance of the student's residence from the school and whether they have siblings already at the school. Another application would also be allocating university students to their rooms in university halls of residence. This problem is similar to the student-project allocation problem because each new student usually selects several residences in order of preference. The only adaption required for this problem is that many students would be assigned to each type of residence. (For a similar problem see the group project model in Ref.~\onlinecite{Kwanashie2015}.) With more work it is reasonable to suggest that our simulated annealing program could be applied to two-sided allocation problems. In particular, it could allocate junior doctors to their first posts at different hospitals. This problem is two-sided because the hospitals have preferences for which doctors they want as well as the doctors having preferences for the hospitals.

\begin{acknowledgments}
The authors have benefited from conversations with Tom Underwood in the early stages of this work.
\end{acknowledgments}

\appendix*
\section{Simulated annealing code for the student-project allocation problem}
\label{sec:appendix}
The input to the available simulated annealing C program\cite{cprog} takes spreadsheet data as input in the form of two comma separated values (CSV) files. The latter format is chosen because it is often useful for the course organizer to collect preferences using one an online survey tool which can output data in CSV form. One of the two input files contains information on the student preferences for projects, and the other provides information regarding the constraints on supervisor workload. In the preferences file, each student (or student pair) is represented by a column and each project by a row. For each student the projects that they have chosen are given an entry of 1 to 4 corresponding to their preferences. Other cells in the row are left blank. In the supervisor constraints file, each row represents a project and each column represents a supervisor. If supervisor $i$ submitted project $j$, the cell will contain a value between 0 and 1, representing the workload the project will require, depending on the nature of the project or whether there are co-supervisors. A feasible solution allows a supervisor to take up to unit workload. For example, a supervisor could supervise two projects with workload 0.5 or one project of 0.5 and one of 0.25. However, a supervisor could not supervise three projects each of workload 0.5.

The program produces a running report on the value of the objective function, allowing the user to monitor how the quality of the allocation improves as the ``temperature'' is reduced. At the end of the annealing schedule, the final allocation is output to a CSV file in the form of (student, allocated project, their rank choice).

Our code uses pseudorandom numbers generated by a separate subroutine. This takes a random seed based on the system time. For efficiency we generate a large number of pseudorandom numbers and store them in an array from which we draw them as required. Once all are used up we replenish the array.

\end{document}